\documentclass[journal]{IEEEtran}

\usepackage{xcolor,soul,framed} 

\colorlet{shadecolor}{yellow}
\usepackage[pdftex]{graphicx}
\graphicspath{{../pdf/}{../jpeg/}}
\DeclareGraphicsExtensions{.pdf,.jpeg,.png}

\usepackage[cmex10]{amsmath}
\usepackage{amssymb}
\usepackage{array}
\usepackage{mdwmath}
\usepackage{mdwtab}
\usepackage{eqparbox}
\usepackage{tabularx}
\usepackage{booktabs}
\usepackage[normalem]{ulem}
\useunder{\uline}{\ul}{}
\usepackage{url}
\usepackage{hyperref}
\hypersetup{
    colorlinks=true,    
    pdfborder={0 0 0},   
    linkcolor=blue,      
    urlcolor=black,       
    citecolor=blue,      
    filecolor=blue       
}
\usepackage{orcidlink}
\usepackage{algorithm}
\usepackage{algorithmic}
\usepackage{multirow}
\usepackage{cleveref}
\usepackage[hang,flushmargin]{footmisc}
\crefname{table}{Table}{Table}
\crefname{figure}{Fig.}{Fig.}
\crefname{equation}{Eq.}{Eq.}
\crefname{section}{Section}{Section}
\crefname{algorithm}{Algorithm.}{Algorithm.}
\newcommand{\nop}[1]{}

\begin{document}
\bstctlcite{}
    \title{Network-wide Freeway Traffic Estimation Using Sparse Sensor Data: A Dirichlet Graph Auto-Encoder Approach}
  \author{Qishen~Zhou$^{\mbox{\orcidlink{0000-0001-8074-4111}}}$,
      Yifan~Zhang$^{\mbox{\orcidlink{0000-0001-8882-4327}}}$,
      Michail~A.~Makridis$^{\mbox{\orcidlink{0000-0001-7462-4674}}}$,~\IEEEmembership{Member,~IEEE,}
      Anastasios~Kouvelas$^{\mbox{\orcidlink{0000-0003-4571-2530}}}$,~\IEEEmembership{Senior Member,~IEEE,}
      Yibing~Wang$^{\mbox{\orcidlink{0000-0001-9937-2055}}}$,~\IEEEmembership{Senior Member,~IEEE,} 
      and~Simon~Hu$^{\mbox{\orcidlink{0000-0002-9832-6679}}}$,~\IEEEmembership{Senior Member,~IEEE}

  \thanks{This work was supported in part by ``Pioneer'' and ``Leading Goose'' R\&D Program of Zhejiang Province (2023C03155), the National Natural Science Foundation of China (52131202, 72350710798), the National Key R\&D Program of China (2023YFB4302600), the Smart Urban Future (SURF) Laboratory, Zhejiang Province, Zhejiang University Sustainable Smart Livable Cities Alliance (SSLCA), and Zhejiang University Global Partnership Fund led by Principal Supervisor Simon Hu, and the Provincial Key R\&D Program of Zhejiang (2024C01180, 2022C01129), the National Natural Science Foundation of China (52272315) led by Co-supervisor Yibing Wang. Qishen Zhou acknowledges support from the China Scholarship Council for his one-year visiting at ETH Zurich. (Corresponding author: Simon Hu.)}
  \thanks{Qishen Zhou, Yibing Wang and Simon Hu are with the Institute of Intelligent Transportation Systems, College of Civil Engineering and Architecture, Zhejiang University, 310058 Hangzhou,  China (e-mail: qishenzhou@intl.zju.edu.cn; wangyibing@zju.edu.cn; simonhu@zju.edu.cn).}
  \thanks{Yifan Zhang is with the Department of Computer Science, City University of Hong Kong (Dongguan), 523808 Dongguan, China (e-mail: yifan.zhang@cityu-dg.edu.cn).}
  \thanks{Michail~A.~Makridis and Anastasios~Kouvelas are with the Institute for Transport Planning and Systems, Swiss Federal Institute of Technology (ETH) Zürich, 8092 Zürich, Switzerland (e-mail:michail.makridis@ivt.baug.ethz.ch; kouvelas@ethz.ch).}}

\markboth{}{}

\maketitle

\begin{abstract}

Network-wide Traffic State Estimation (TSE), which aims to infer a complete image of network traffic states with sparsely deployed sensors, plays a vital role in intelligent transportation systems. With the development of data-driven methods, traffic dynamics modeling has advanced significantly. However, TSE poses fundamental challenges for data-driven approaches, since historical patterns cannot be learned locally at sensor-free segments. Although inductive graph learning shows promise in estimating states at locations without sensor, existing methods typically handle unobserved locations by filling them with zeros, introducing bias to the sensitive graph message propagation. The recently proposed Dirichlet Energy-based Feature Propagation (DEFP) method achieves State-Of-The-Art (SOTA) performance in unobserved node classification by eliminating the need for zero-filling. However, applying it to TSE faces three key challenges: inability to handle directed traffic networks, strong assumptions in traffic spatial correlation modeling, and overlooks distinct propagation rules of different patterns (e.g., congestion and free flow). We propose DGAE, a novel inductive graph representation model that addresses these challenges through theoretically derived DEFP for Directed graph (DEFP4D), enhanced spatial representation learning via DEFP4D-guided latent space encoding, and physics-guided propagation mechanisms that separately handles congested and free-flow patterns. Experiments on three traffic datasets demonstrate that DGAE outperforms existing SOTA methods and exhibits strong cross-city transferability. Furthermore, DEFP4D can serve as a standalone lightweight solution, showing superior performance under extremely sparse sensor conditions.

\end{abstract}


%

\section{Introduction}

\IEEEPARstart{S}{tate} estimation is concerned with reconciling noisy observations of a physical system with its mathematical model for inferring unmeasurable states and denoising measurable ones \cite{courseStateEstimationPhysical2023}. In transportation systems, it specifically refers to the process of inferring traffic state variables (e.g., flow, density and speed) at sensor-free road segments using traffic data from sparsely sensor-equipped ones. Traffic State Estimation (TSE) serves as a crucial component in freeway traffic control and management, as traffic sensors cannot be installed ubiquitously due to technological and financial constraints \cite{seoTrafficStateEstimation2017}. 

TSE has been widely considered as a challenging task since there is no historical data at unobserved segments for model calibration. Over the past decade, the transportation community has witnessed the advances in physical model-driven approaches for TSE \cite{saeedmanesh2021ExtendedKalmanFilter, WANG2022EKF,MAKRIDIS2023-UKF,trinh2024StochasticSwitchingMode}, which have provided rigorous theoretical frameworks for analyzing and predicting traffic dynamics. Despite their success, these approaches suffer from the lack of systematic hyper-parameter tuning mechanisms and require substantial domain expertise. Furthermore, their performance is inherently limited by the various assumptions made during the mathematical formulation of traffic dynamics. With the unprecedented growth of computing power and availability of massive datasets, machine learning and deep learning techniques have achieved remarkable breakthroughs and widespread adoption across various domains given their capability of modeling the complex non-linear relationship. Following this trend, data-driven approaches for TSE have gained increasing attention in recent years as promising alternatives to overcome the limitations of traditional physical model-based methods.

Data-driven TSE methods can be broadly classified into two paradigms: transductive and inductive approaches. Transductive methods, exemplified by tensor completion \cite{Lei2022Kriging,nie2023letc} and node embedding \cite{xu2020GEGANNovelDeep}, require complete model retraining in response to even the slightest changes in sensor configurations or new unobserved locations emerge. This leads to substantial computational overhead and reduced operational flexibility. In contrast, inductive methods can seamlessly generalize to new unobserved locations of interest, handle sensor configurations variations arising from inevitable sensor retirement and new installations. Moreover, inductive methods enable cross-region model deployment without retraining, making them particularly valuable for real-world traffic monitoring systems.

Traditional inductive models, represented by Gaussian Process \cite{Rasmussen2005Nov,cressie2011statistics}, rely on flexible kernel functions to capture spatial correlations, yet face computational limitations that restrict their applicability to large-scale networks. The majority of modern inductive methods are built upon Graph Neural Networks (GNN). Leveraging a powerful message passing mechanism \cite{gilmer2017MPNN} to effectively capture and abstract shared spatial dependencies between graph nodes, GNN enables inference of unobserved node states from observed ones. This generalizable spatial learning capability makes GNNs particularly suitable for TSE tasks, as locations of interest can be naturally modeled as graph nodes with edges reflecting the physical road network. Through careful design of network architectures and training strategies, inductive GNN-based methods have demonstrated promising estimation performance across various traffic datasets \cite{wu2021IGNNK,Xu2023AGNPNetworkwideShortterm,xu2023kits, nieBetterTrafficVolume2023}. 

Despite these advances, one critical issue worth noting is that existing inductive GNN-based methods commonly fill unobserved node features with zeros. While zero-filling has been adopted as a seemingly reasonable solution for missing data scenarios across various machine learning tasks, as it eliminates the influence of missing features by nullifying their associated weight updates during training, many existing studies have shown that it would introduce bias \cite{pengMultiviewGraphImputation2024} and lead to performance degradation \cite{yiWhyNotUse2019}. To address this issue, \cite{rossi2022FP} proposed an efficient Dirichlet Energy-based Feature Propagation algorithm (DEFP) for preliminary node attribute filling in social and citation networks, demonstrating its effectiveness on downstream node classification tasks even with a large proportion of attribute-missing nodes. However, directly applying DEFP to TSE tasks faces three major challenges. First, DEFP is currently limited to undirected graphs, while traffic patterns exhibit clear directionality in their propagation. Second, its core mechanism of minimizing Dirichlet energy essentially enforces spatial smoothness of features across neighboring nodes. While this smoothness assumption works well for citation networks (e.g., Cora, CiteSeer, PubMed), directly applying it to traffic spatiotemporal data in the original feature space may not adequately capture the complex spatial relationships inherent in traffic patterns. For instance, traffic states between adjacent roads can exhibit abrupt changes during peak hours or incidents, violating the basic smoothness assumption. Third, this method processes all patterns in the data uniformly during feature propagation, while in traffic dynamics, different patterns follow distinct physical rules. Specifically, congested patterns propagate upstream at slower and more volatile speeds, whereas free-flow patterns spread downstream more rapidly and stably \cite{lighthill1955kinematic}. These challenges indicate the need for a more sophisticated approach to handle directed graphs, capture complex spatial dependencies, and incorporate traffic flow physics into the feature propagation mechanism.

To address these challenges, we propose Dirichlet Graph Auto-Encoder (DGAE), a novel inductive graph representation model that advances the current data-driven state-of-the-art in TSE. The main contributions of this paper can be summarized as follows:

\begin{itemize}
    
    \item We theoretically derive Dirichlet Energy-based Feature Propagation for Directed graph (DEFP4D), enabling more accurate feature propagation in modeling of traffic dynamics where directions are critical.
    
    \item We create a mutually reinforcing mechanism by applying DEFP4D in a learnable latent space that constructed by graph auto-encoder, where DEFP4D guides the learning of spatially smooth latent feature representations, while the optimized latent space provides an enhanced workspace for feature propagation, jointly improving the estimation for unobserved nodes.
    
    \item We develop a physics-guided DEFP4D based on the flexible foundation of graph auto-encoder, which separately propagates congested and free-flow signals to better characterize traffic flow dynamics.

    \item We conduct comprehensive experiments on three open-source traffic datasets, demonstrating that DGAE reduces estimation errors by up to 10.53\% while requiring 14.5\% fewer sensors compared to state-of-the-art baselines. The model also exhibits strong generalization capabilities in challenging cross-city transfer scenarios.
\end{itemize}

Additionally, we note that DEFP4D can be directly applied to the original feature space, serving as a lightweight variant of DGAE. This variant not only demonstrates competitive performance comparable to existing inductive graph representation learning baselines, but also surpasses DGAE when deployed sensors are extremely sparse. These findings provide flexible solutions for various practical deployment scenarios.

The remainder of the paper is organized as follows: \cref{sec:LR} delivers a literature review related to data-driven traffic state estimation. \cref{sec:PF} provides the formalization of the problem. \cref{sec:Method} details the Dirichlet graph auto-encoder. \cref{sec:nm} reports the performance and ablation results of the proposed method. \cref{sec:diss} discusses the limitations of the current work and outlines future research directions. \cref{sec:con} concludes the paper.

\section{Literature review} \label{sec:LR}

Traffic State Estimation (TSE) methods are categorized into three branches: physical model-driven, data-driven, and hybrid approaches that combine the former two. Following our research focus on addressing the limitations in data-driven methods, this section primarily reviews recent advances in data-driven TSE. We refer interested readers to the comprehensive reviews in \cite{seoTrafficStateEstimation2017}, \cite{WANG2022EKF}  and \cite{hu2022high} of physical model-driven approaches. For the hybrid approach, recent methods \cite{Shi2021PIDL,Shi2022PIDL,YUAN2021PIGP,Yuan2022PIGP,zhangPhysicsinformedDeepLearning2024, wang2024KnowledgedataFusionOriented} follow the Physics-Informed Neural Network (PINN) framework \cite{RAISSI2019PINN}, which integrates physical priors of traffic flow in the form of partial differential equations \cite{lighthill1955kinematic,richards1956shock,aw2000resurrection,zhang2002non} into network training to combine data-driven and physical model advantages. However, these methods show limited generalization beyond spatio-temporal boundary of training data \cite{Kim2021DPM}, making online state estimation challenging. For detailed discussions on this topic, we direct readers to the comprehensive reviews by \cite{karniadakisPhysicsinformedMachineLearning2021} and \cite{thodiFourierNeuralOperator2024}. In the following subsections, we review the recent developments in transductive and inductive data-driven methods respectively. Before proceeding with the detailed review, it is important to note that spatio-temporal data kriging problems \cite{Lei2022Kriging,wu2021IGNNK}, which aim to predict values at unobserved locations by exploiting spatial correlations, fundamentally align with the scope of traffic state estimation. Therefore, we include these kriging-based approaches as part of the traffic state estimation methodology in this review.

\subsection{Transductive data-driven traffic state estimation}

Transductive data-driven TSE methods primarily focus on tensor completion approaches, which aim to recover missing values in partially observed tensors using low-rank hypothesis \cite{song2019TensorCompletionAlgorithms}. However, in TSE, traffic data tensors always have entire rows or columns missing due to limited sensor coverage, making low-rank assumptions alone insufficient to guarantee solution uniqueness \cite{nie2023letc}. Thus, modern tensor completion for TSE typically incorporates additional spatio-temporal regularization terms \cite{Lei2022Kriging}. For spatial modeling, most existing methods employ graph Laplacian regularization \cite{Bahadori2014KrigingGLTL, zhangNetworkwideTrafficFlow2020, nie2023letc} or Gaussian processes \cite{zhou2012KPMF, Lei2022Kriging}. While graph Laplacian regularization is relatively easy to implement, it faces challenges in hyperparameter tuning especially when defining kernel-based graph edge weights; Gaussian processes offer strong modeling capacity but at the cost of high computational complexity. For temporal modeling, methods have evolved from local Markov \cite{xiong2010cfiltering} and autoregressive regularization \cite{yu2016TemporalRegularizedMatrix} that primarily capture short-term dependencies to approaches that model both short-term continuity and long-term periodicity \cite{nie2023letc}. Besides tensor completion methods, there are also attempts combining generative models \cite{goodfellow2020generative} with node embeddings \cite{yan2007GraphEmbeddingExtensions} enables accurate inference of unobserved location data based on sampled observed neighbors \cite{xu2020GEGANNovelDeep}, though this direction remains relatively unexplored in the literature.

Despite the promising performance of these transductive methods, they require retraining when sensor configurations change due to retirement or new installations, and cannot generalize to new unobserved locations, revealing their inherent fragility in real-world applications.

\subsection{Inductive data-driven traffic state estimation}

Inductive methods exhibit notable advantages over transductive approaches through their ability to generalize to unobserved locations, robustness against sensor network, and enable cross-regional deployment without retraining, making them valuable for operational monitoring systems. In early research, Kriging methods \cite{saito2005OKriging} served as the primary approach for inductive data-driven traffic state estimation. These methods rely on variogram functions to model the relationship between spatial distance and state correlation, thereby inferring the states of unobserved nodes. The performance of Kriging methods heavily depends on the choice of variogram functions, and being essentially equivalent to Gaussian processes, they are characterized by high computational costs. 

With the advent of deep learning, inductive graph representation learning methods have gained significant attention. These approaches leverage the powerful expressiveness of Graph Neural Networks (GNNs) to capture complex spatio-temporal dependencies through message-passing mechanisms. For instance, \cite{wu2021IGNNK} proposed an inductive graph spatio-temporal Kriging method that outperforms Kriging and tensor completion approaches. In their method, random subgraph construction and node masking were introduced during training to enhance model generalization across varying sensor configurations. However, this method only utilized static graphs, overlooking potential temporal variations in spatial relationships. Subsequent studies addressed this limitation by either coupling multiple spatial aggregation operators \cite{wu2021SATC,zhou2025MoGERNNInductiveTraffic} or constructing multi-graph networks based on multi-source data \cite{Zheng2023Increase}. Additionally, some researchers argued that the training strategy proposed by \cite{wu2021IGNNK} leads to fewer available sensors and reduced graph size during training compared to testing, resulting in information loss. Therefore, they proposed improvements such as prediction task assistance \cite{WEI2024IAGCN} and incremental training \cite{xu2023kits}. Despite these advancements, existing works commonly handle missing values by simply filling them with zeros, which tends to introduce noise into message propagation in GNN and thus compromise model performance \cite{yiWhyNotUse2019,rossi2022FP}. Beyond GNN architectures, recent work has adopted conditional diffusion models for traffic state estimation \cite{lei2024ConditionalDiffusionModel}, though the well-known issue of lengthy inference times remains unaddressed, potentially limiting their practical deployment.

\section{Problem formulation} \label{sec:PF}

Consider a traffic network with \(N\) locations of interest (hereafter locations and nodes are used interchangeably), \(\mathcal{V} = \{\nu_1, \nu_2, \ldots, \nu_N\}\), where \(N^o\) nodes (\(N^o < N\)) are equipped with traffic sensors. The network is formulated as a directed graph \(\mathcal{G} = (\mathcal{V}, \mathbf{A})\), with \(\mathbf{A} \in \mathbb{R}^{N \times N}\) representing the adjacency matrix. Following previous studies \cite{li2018dcrnn}, we construct \(\mathbf{A}\) using Gaussian kernel of the pairwise distances as edge weights:
\begin{equation}
A_{i j}= \begin{cases}\operatorname{exp}\left(-\frac{d_{ij}^2}{\sigma^2}\right), & \text{if there's a route from } \nu_i \text{ to } \nu_j  \\ 0, & \text{otherwise} \end{cases}
\end{equation}
here, \(d_{ij}=\operatorname{dist}(\nu_i,\nu_j)\) represents the travel distance from node \(\nu_i\) to \(\nu_j\), while \(\sigma\) and \(\kappa\) are the standard deviation of distances and the distance threshold, respectively. The goal of this work is to estimate the speeds of \(N^u\) unmonitored locations ($N^u = N - N^o$) based on the measurement speeds from $N^o$ observed locations and the topological structure $\mathcal{G}$:
\begin{equation}
    f_{\theta} (\mathbf{X}^o, \mathcal{G}) = {\mathbf{X}^u} 
\end{equation}
where $\mathbf{X}^o \in \mathbb{R}^{N^o \times L}$ denotes the temporal sequences collected from $N^o$ observed locations over a window of length $L$, and $\mathbf{X}^u \in \mathbb{R}^{N^u \times L}$ represents the speed to be estimated at $N^u$ unmonitored locations during the same time period. 

\begin{figure*}
    \centering
    \includegraphics[width=0.8\textwidth]{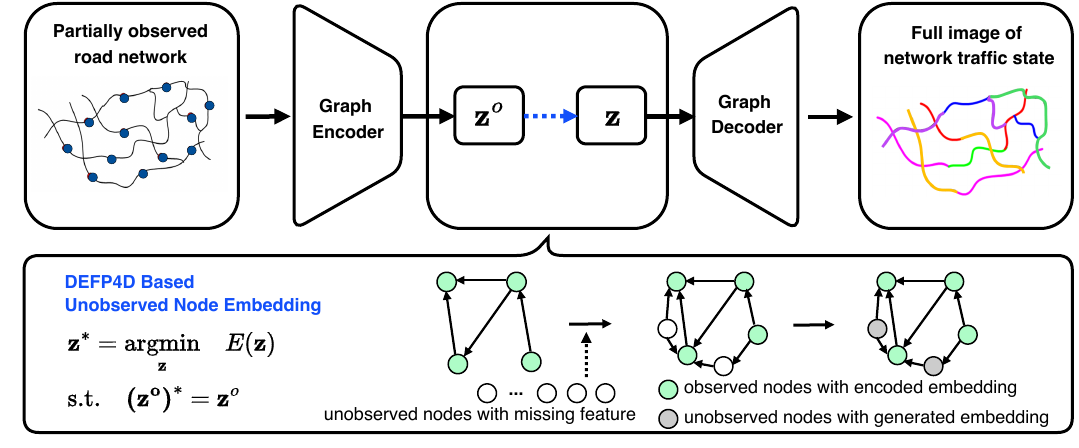}
    \vspace{-3mm}
    \caption {Illustration of foundational framework Dirichlet graph auto-encoder. 
    }
    \vspace{-12pt}
    \label{fig:DGAE-1}
\end{figure*}

\section{Methodology} \label{sec:Method}
In this section, we introduce our novel Dirichlet Graph Auto-encoder (DGAE) model, with its overall architecture illustrated in \cref{fig:DGAE-1}. The architecture consists of two major components: the Graph Auto-Encoder (GAE) structure and the unobserved node embedding generation based on Dirichlet Energy-based Feature Propagation For Directed graph (DEFP4D). DGAE first takes partially observed road network data as input, which only contains observed nodes (blue dots). Through the graph encoder, these observed data are encoded into initial node embeddings $z_o$, which are then extended to complete embeddings z that include all nodes based on DEFP4D. Finally, the graph decoder decodes the complete embeddings to infer the global traffic state for all nodes (road segments) of interest. Given that DEFP4D serves as the cornerstone of DGAE by providing high-quality embeddings for unobserved nodes, in the following subsections, we first elaborate the DEFP4D, followed by the basic framework of DGAE, and finally present the refined DGAE model from a physics-guided perspective. In addition, to guide the model in inferring unobserved node states from observed nodes, a dynamic masking-based training strategy is introduced in the last subsection.

\subsection{DEFP4D:Dirichlet energy-based feature propagation for directed graph} \label{sec:fp}
The Dirichlet energy is a metric for measuring the spatial smoothness of signals over graphs. For signals at graph nodes $\mathbf{x}$, it is defined as \(E(\mathbf{x}) = \frac{1}{2}\sum_{i,j}A_{i,j}|x_i-x_j|^2\), which can be expressed in matrix form as:
\begin{equation}
E(\mathbf{x}) = \frac{1}{2}\mathbf{x}^\top(\mathbf{D}_o + \mathbf{D}_I - 2\mathbf{A})\mathbf{x}
\end{equation}
where $\mathbf{D}_o$ and $\mathbf{D}_I$ are the out-degree and in-degree matrices respectively. For undirected graphs, where $\mathbf{A} = \mathbf{A}^\top$, this energy can be elegantly expressed using the graph Laplacian operator:
\begin{equation}
E(\mathbf{x}) = \mathbf{x}^\top\mathbf{L}\mathbf{x}
\end{equation}
where $\mathbf{L} = \mathbf{D} - \mathbf{A}$ is the graph Laplacian matrix, with $\mathbf{D}$ being the degree matrix.

Dirichlet energy has been widely employed to exploit the spatial dependencies among nodes. This principle has been particularly influential in tensor completion-based TSE algorithms \cite{zhangNetworkwideTrafficFlow2020,Lei2022Kriging,nie2023letc}. Research has also shown that lots of Graph Neural Networks (GNNs) are essentially equivalent to solving optimization problems regularized by Dirichlet energy \cite{zhuInterpretingUnifyingGraph2021}. Given its ability to quantify spatial smoothness on graphs, it has been utilized to control the over-smoothing issue that GNNs are frequently criticized for \cite{zhouDirichletEnergyConstrained2021}. \cite{rossi2022FP} proposed Dirichlet Energy based graph node Feature Propagation (DEFP) algorithm for preliminary feature imputation of missing nodes on the undirected graph. After this initial filling through DEFP, subsequent GNN modules achieved remarkable performance on downstream tasks. Drawing from these insights, we developed an enhanced variant of DEFP optimized For Directed graph (DEFP4D). The core idea of DEFP4D is formulated as an optimization problem that minimizes the Dirichlet energy while preserving the observed signal values. Specifically, it seeks to estimate the signals of unobserved nodes on the directed graph by solving:
\begin{equation}
\begin{aligned}
    \min_{{\mathbf{x}}} \quad & E({\mathbf{x}}) = \frac{1}{2}\mathbf{x}^\top(\mathbf{D}_o + \mathbf{D}_I - 2\mathbf{A})\mathbf{x} \\
    \text{s.t.} \quad & {x}_i = x^o_i, \text{if } i \in \mathcal{O}
\end{aligned}
\end{equation}
where \(\mathcal{O}\) denotes the set of observed nodes.

To solve this optimization problem, we consider its associated gradient flow \(\dot{\mathbf{x}}(k) = -\nabla E(\mathbf{x}(k)) \) with Boundary conditions (BC) and Initial Conditions (IC). The solution for unobserved nodes state estimation can be obtained by taking the limit as \(\hat{\mathbf{x}}^u = \lim_{k \rightarrow \infty} \mathbf{x}^u(k)\): 
\begin{equation}
    \begin{aligned}
        \dot{\mathbf{x}}(k)=-\mathbf{P} \mathbf{x}(k) \quad [BC]: \mathbf{x}^o(t)=\mathbf{x}^o \quad [IC]: \mathbf{x}(0)=\left[\begin{array}{c}
        \mathbf{x}^o \\
        \mathbf{x}^u(0)
        \end{array}\right] 
    \end{aligned}
    \label{eq:problem}
\end{equation}
where \(\mathbf{P}=(\mathbf{D}_o+\mathbf{D}_I)-(\mathbf{A}+\mathbf{A}^\top)\). This formulation corresponds to a dynamical system with Dirichlet boundary conditions, where the solution values are fixed at the observed nodes. The matrix form of this problem is as follows:
\begin{equation}
    \left[\begin{array}{c}
    \dot{\mathbf{x}}^o(k) \\
    \dot{\mathbf{x}}^u(k)
    \end{array}\right] = -\left[\begin{array}{cc}
    \mathbf{0} & \mathbf{0} \\
    \mathbf{P}^{uo} & \mathbf{P}^{uu}
    \end{array}\right]\left[\begin{array}{c}
    \mathbf{x}^o \\
    \mathbf{x}^u(k)
    \end{array}\right]
\end{equation}
where \(\mathbf{P}^{uo}\) is the block matrix capturing the interactions between unobserved and observed nodes, and \(\mathbf{P}^{uu}\) is the block matrix describing the interactions within unobserved nodes. 

To derive a practical algorithm, we discretize the continuous-time dynamics using forward Euler scheme:
\begin{equation}
    \left[\begin{array}{c}
    \mathbf{x}^o(k+1) -\mathbf{x}^o(k)\\
    \mathbf{x}^o(k+1)-\mathbf{x}^o(k)
    \end{array}\right] = -\mathbf{\Delta} \left[\begin{array}{cc}
    \mathbf{0} & \mathbf{0} \\
    \mathbf{P}^{uo} & \mathbf{P}^{uu}
    \end{array}\right]\left[\begin{array}{c}
    \mathbf{x}^o \\
    \mathbf{x}^u(k)
    \end{array}\right]
\end{equation}

With a choice of time step $\mathbf{\Delta} = (\mathbf{D_o}+\mathbf{D_I})^{-1}$, we obtain a concise update rule as:
\begin{equation}
    \mathbf{x}(k+1) =\left[\begin{array}{cc}
    \mathbf{I} & \mathbf{0} \\
    \tilde{\mathbf{A}}^{uo} & \tilde{\mathbf{A}}^{uu}
    \end{array}\right]\mathbf{x}(k) 
    \label{eq:update}
\end{equation}
where \(\tilde{\mathbf{A}}=(\mathbf{A+A^\top})/(\mathbf{D_o}+\mathbf{D_I})\). Notably, when specializing to undirected graphs where the Dirichlet energy takes the form of \(\frac{1}{2}\sum w_{i,j}(x_j-x_j)^2\), \(w_{i,j}\) denotes the entries of the normalized adjacency matrix \(D^{(-1/2)}AD^{(-1/2)}\), \(\tilde{\mathbf{A}}\) in \cref{eq:update} precisely reduces to \(D^{(-1/2)}AD^{(-1/2)}\), with \(\Delta = \mathbf{I}\), which coincides with the well-known diffusion kernel in graph convolution neural network \cite{kipf2017semisupervised}.

Inspired by \cite{rossi2022FP}, this update rule can be interpreted as an iterative process where node features \(\mathbf{x}\) are repeatedly multiplied by the transition matrix \(\mathbf{\tilde{A}}\), while maintaining the observed node features fixed at their original values in each iteration, as detailed in \cref{alg}. In our experiments across three test datasets, the imputation of unobserved node features in the original feature space converges rapidly, typically within 90 iterations.

\begin{algorithm}[t]
\caption{Gradient descent for Dirichlet energy minimization with Dirichlet boundary condition}
\label{alg}
\begin{algorithmic}[1]
\REQUIRE Feature vector \( \mathbf{x}^o \) for observed nodes, transition matrix \( \tilde{\mathbf{A}} \), number of iterations \( K \).
\ENSURE Updated node feature vector \( \mathbf{x}(K) \)
\STATE $\mathbf{c} \leftarrow \mathbf{x}^o$
\STATE Randomly initialize unobserved nodes feature \(\mathbf{x}^u(0)\).
\FOR{\( k = 1 \) to \( K\)}
    \STATE Update node features: $\mathbf{x}(k+1)\leftarrow \tilde{\mathbf{A}} \mathbf{x}(k)$
    \STATE Reset observed node features: $\mathbf{x}^o(k+1) \leftarrow \mathbf{c}$
\ENDFOR
\end{algorithmic}
\end{algorithm}

\subsection{DGAE:Foundational framework of Dirichlet graph auto-encoder} \label{sec:ae}

Building upon the DEFP4D introduced in the previous subsection, we begin with the GAE component to present our DGAE model. GAEs have emerged as a powerful tool for learning node feature embeddings and graph structural representations. By encoding both node features and graph structural information into a latent space, they enable effective node embeddings that benefit downstream tasks such as classification and prediction. The basic architecture consists of an encoder and a decoder, formulated as:
\begin{equation}
\mathbf{z} \in \mathbb{R}^{N \times D} = \mathcal{G}^e_{\theta}(\mathbf{x} \in \mathbb{R}^{N \times L}, \mathbf{A} \in \mathbb{R}^{N \times N})
\label{eq:tge}
\end{equation}
\vspace{-16pt}
\begin{equation}
\mathbf{y} \in \mathbb{R}^{N \times L} = \mathcal{G}^d_{\theta}(\mathbf{z} \in \mathbb{R}^{N \times D}, \mathbf{A} \in \mathbb{R}^{N \times N})
\label{eq:tgd}
\end{equation}
where \(N\) denotes the number of nodes in the considered graph, \(D\) represents the dimensions of latent representations. \(L\) denotes the sequence length of the feature. \(\mathcal{G}^e_{\theta}\) and \(\mathcal{G}^d_{\theta}\) denote the graph encoder and decoder operations.

From the formulations above, we can observe that GAE requires a complete input feature matrix \(\mathbf{x} \in \mathbb{R}^{N \times L}\) for all nodes to perform encoding. However, in state estimation tasks, we often encounter scenarios where features are only partially observed, resulting in an incomplete \(\mathbf{x}\). Therefore, when applying GAE to state estimation tasks, preliminary feature filling for unobserved nodes becomes necessary, commonly achieved through zero-filling. However, as discussed in the introduction, this approach inevitably introduces substantial noise. While \cite{rossi2022FP} proposed DEFP-filling as an effective strategy for feature imputation in social and citation networks, its suitability for traffic spatiotemporal data is questionable. As discussed earlier, DEFP minimizes Dirichlet energy, enforcing feature smoothness across neighboring nodes, which might be ill-suited for original feature space of traffic data.

To address this fundamental issue, the proposed DGAE framework adopts a two-stage strategy to generate node embeddings: first, it encodes the observed nodes (shown as blue dots in \cref{fig:DGAE-1}) extracted from the complete road network; subsequently, based on the embeddings of observed nodes, it generates embeddings for unobserved nodes in the latent space through the DEFP4D mechanism, as illustrated in the lower sub-figure of \cref{fig:DGAE-1}:
\begin{equation}
\mathbf{z}^o \in \mathbb{R}^{N^o \times D} = \mathcal{G}^e_{\theta}(\mathbf{x} \in \mathbb{R}^{N^o \times L}, \mathbf{A} \in \mathbb{R}^{N^o \times N^o})
\label{encoder}
\end{equation}
\vspace{-16pt}
\begin{equation}
\mathbf{z} \in \mathbb{R}^{N \times D} = \operatorname{DEFP4D}(\mathbf{z}^o \in \mathbb{R}^{N^o \times D}, \mathbf{A} \in \mathbb{R}^{N \times N})
\end{equation}
where $\mathbf{z}^o$ represents the embeddings of observed nodes. After get the embeddings of all nodes, they are fed into the graph decoder for global state estimation, consistent with the original GAE framework.

\begin{figure}
    \centering
    \includegraphics[width=1.0\columnwidth]{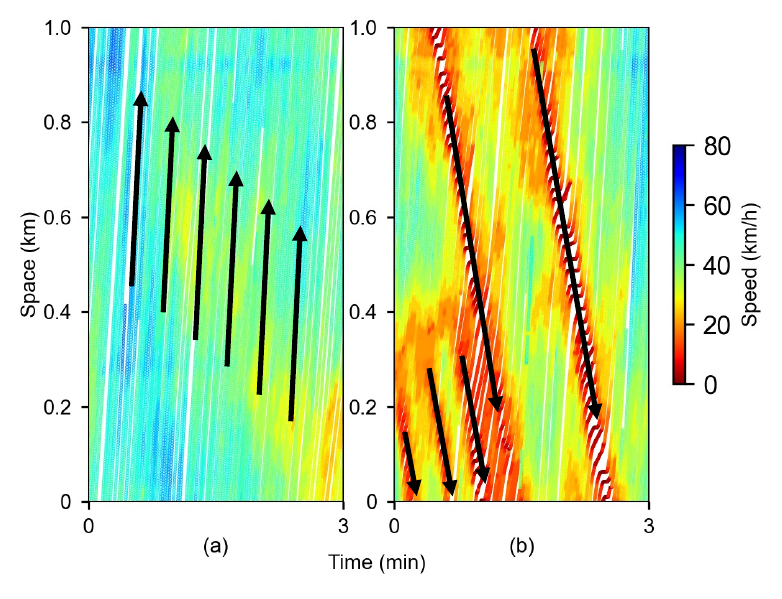}
    \vspace{-10mm}
    \caption{Illustration of bidirectional diffusion dynamics of traffic flow: an evidence from U.S. Highway 101.}
    \vspace{-12pt}
    \label{fig:bi-dyn}
\end{figure}

\begin{figure*}
    \centering
    \includegraphics[width=0.9\linewidth]{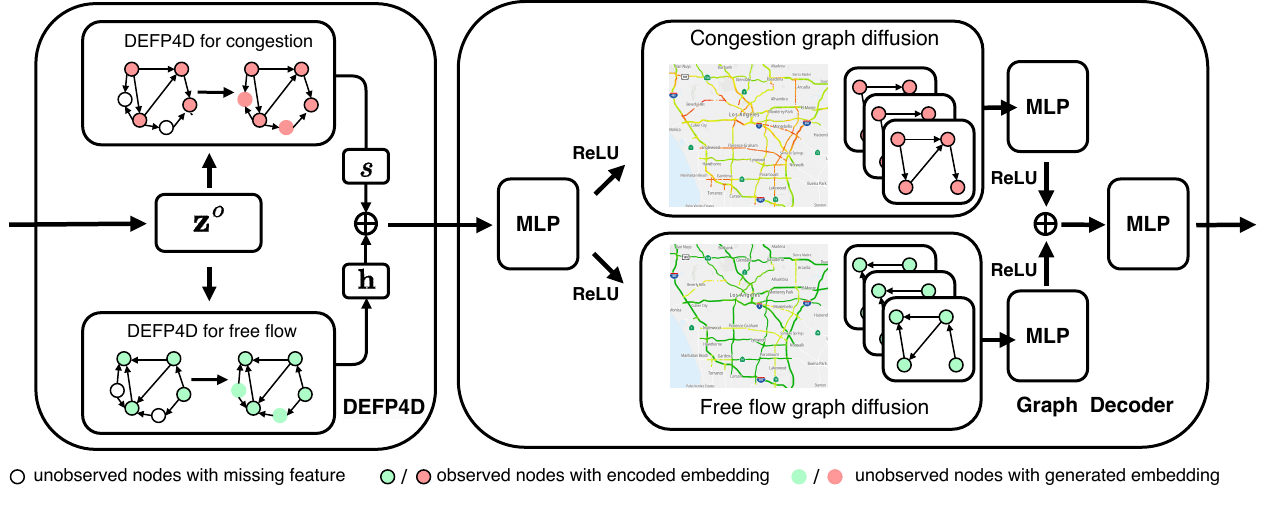}
    \vspace{-5mm}
    \caption {Illustration of the refined DGAE architecture. The Graph Encoder structure, identical to the Graph Decoder, is omitted here for brevity. MLP denotes Multi-Layer Perceptron, ReLU denotes the activation function and $\oplus$ represents element-wise addition operation.}
    \vspace{-12pt}
    \label{fig:DGAE-2}
\end{figure*}

Applying DEFP4D in the latent space creates a mutually reinforcing mechanism: DEFP4D guides the learning of spatially smooth feature representations, while the optimized latent space provides an enhanced workspace for feature propagation, jointly improving the quality for observed and unobserved node embeddings. Furthermore, leveraging the flexibility of GAE, we incorporate auxiliary features to enhance node embeddings. Specifically, we concatenate the original node features with time stamp embedding \(\mathbf{E}_t \in \mathbb{R}^{N^o \times D_t}\) and missing feature position embedding \(\mathbf{E}_m \in \mathbb{R}^{N^o \times D_m}\) to enhance temporal modeling and improve the model's robustness against temporarily missing features. Based on this, the \(\mathbf{x}\) in \cref{encoder} is replaced by:
\begin{equation}
    \mathbf{x} \in \mathbb{R}^{N^o \times (L + D_t+D_m)} = (\mathbf{x},\mathbf{E}_t,\mathbf{E}_m)
\end{equation}
where \(\mathbf{E}_t\) is a learnable identity embedding representing the timestamp of the last element in the input sequence, and \(\mathbf{E}_m\) is derived by transforming a binary mask matrix (which has the same shape as original \(\mathbf{x}\) and indicates the presence or absence of each element) through a Multi-Layer Perceptron (MLP) network.

As for the graph encoder and decoder, they are sharing identical structures, with graph diffusion convolution \cite{gasteigerDiffusionImprovesGraph2019} as their core module. A graph diffusion convolution operation is performed by multiplying the diffusion matrix \(\mathbf{S}\) with the input features \(\mathbf{x}\). The general form of diffusion matrix \(\mathbf{S}\) is as follows:
\begin{equation}
    \mathbf{S} = \sum_{k=0}^\infty\theta_k\mathbf{T}^k
    \label{eq:diffusion}
\end{equation}
where \(\mathbf{T}\) denotes the transition matrix, \(\theta_k\) represents the weighting coefficients and \(k\) indicates the diffusion step that determines how far the information can propagate through the graph. The detailed structure of the graph encoder and decoder will be presented in the next subsection. We combine MLPs with graph diffusion convolution to construct the graph encoder and decoder, and the detailed architecture will be presented in the next subsection.

To better align with TSE characteristics of inferring node states through neighboring features, we modify the diffusion matrix by setting \(k\) to start from 1 rather than 0, as k=0 corresponds to self-features. Typically, transition matrix can be \(\mathbf{D}_o^{-1}\mathbf{A}\) and \(\mathbf{D}^{(-1/2)}\mathbf{A} \mathbf{D}^{(-1/2)}\) in directed graph and undirected graph, respectively. In this work, we adopt \(\tilde{\mathbf{A}}\) derived from \cref{eq:update} as the transition matrix to maintain consistency with the idea of minimize the Dirichlet energy of the graph. Following \cite{chung2007HeatKernelPagerank}, we set the weighting coefficients \(\theta_k=\alpha(1-\alpha)^k\), where \(\alpha\) controls the decay rate of diffusion influence.

\subsection{Dirichlet graph auto-encoder guided by the physics knowledge}
Traffic flow propagation not only exhibits directionality, but different patterns also manifest distinct propagation directions and dynamic behaviors. As illustrated in \cref{fig:bi-dyn}, we visualize the spatiotemporal evolution of velocity fields based on the NGSIM \cite{kovvali2007VideoBasedVehicleTrajectory} vehicle trajectory dataset from US Highway 101. The color bar indicates speed magnitude (km/h), where red represents low speed and blue represents high speed. Subplots (a) and (b) demonstrate free-flow and congested states respectively: under free-flow conditions, traffic patterns propagate downstream along the traffic direction; under congested conditions, the shock waves \cite{lighthill1955kinematic} separating arrival flow and congested flow propagate upstream against the traffic direction. These two patterns not only differ in propagation direction but also show significant differences in propagation speed: the free-flow propagation speed is typically faster and more stable, while the congestion shock wave propagates more slowly and is more susceptible to local perturbations. This difference in propagation speeds directly affects the spatial continuity characteristics of traffic states, resulting in distinct spatial smoothness properties between the two patterns. Therefore, we argue that when applying DEFP4D for feature propagation, these two types of signals should be treated separately.

Based on this observation, we propose a physics-informed DEFP4D that decouple the signals into congestion signals and free-flow signals when doing the features propagation. Specifically, we decompose the transition matrix \(\tilde{\mathbf{A}}\) derived in \cref{eq:update} into \(\tilde{\mathbf{A}}_{\text{cog}}\) and \(\tilde{\mathbf{A}}_\text{free}\) :
\begin{equation}
    \tilde{\mathbf{A}}_{\text{cog}}  = \frac{\mathbf{A}}{\mathbf{D_o}+\mathbf{D_I}}
\end{equation}
\begin{equation}
    \tilde{\mathbf{A}}_\text{free}  = \frac{\mathbf{A}^\top}{\mathbf{D_o}+\mathbf{D_I}}
\end{equation}

\begin{table*}[t]
\caption{Summary of datasets.}
\vspace{-2mm}
\label{tab:dataset}
\centering
\setlength{\tabcolsep}{0.3\tabcolsep}  
\begin{tabular*}{\textwidth}{l@{\extracolsep{\fill}}cccc}
\toprule
Dataset   & Time Period              & Sensors Number & Sampling Frequency & Missing Rate \\ \midrule
METR-LA   & 01/03/2012 - 30/06/2012 & 207           & 5 min             & 8.11\%       \\
PEMS-BAY  & 01/01/2017 - 31/05/2017 & 325           & 5 min             & 0.00\%       \\ 
PEMSD7(M) & 01/05/2012 - 30/06/2012 & 228           & 5 min             & 0.00\%       \\
\bottomrule
\end{tabular*}
\end{table*}

These two transition matrices that are responsible for congestion and free-flow signal propagation are also applied in the design of graph diffusion convolution matrices in \cref{eq:diffusion}, resulting in congestion diffusion matrix \(\mathbf{S}_{\text{cog}}\) and free-flow diffusion matrix \(\mathbf{S}_\text{free}\), which jointly control the propagation directions of features in the spatiotemporal graph.

Based on the above physical insights about traffic flow patterns, the detailed structure of the refined DGAE is shown in \cref{fig:DGAE-2}. Since the graph encoder and decoder share identical structures, both consisting of graph diffusion convolution layers and MLP layers, we omit the graph encoder in the figure for brevity.

\subsection{Dynamic masking-based model training}

Due to the absence of historical data for unobserved nodes in the TSE task, we develop a training strategy to guide the model in inferring unobserved node states from observed nodes. During training, we randomly mask a subset of observed nodes, whose features are not fed into the model but are only used to evaluate prediction errors, thereby simulating the application environment and guiding model training through backpropagation. In addition, the masked nodes are randomly resampled in each training batch to enhance the model's generalization capability across different unobserved node distributions.

\section{Numerical experiments} \label{sec:nm}
In this section, several experiments are set to answer the following question:

\begin{itemize}
    \item \textbf{RQ1}: State estimation performance (\cref{sec:q1}). How do the proposed DGAE and its degenerate version DEFP4D perform in traffic state estimation at sensor-free locations compared to other benchmark methods?
    
    \item \textbf{RQ2}: Impact of sensor deployment density (\cref{sec:q2}). How does the estimation performance of DGAE and DEFP4D at unobserved locations vary with different levels of sensor deployment density?
    
    
    
    \item \textbf{RQ3}: Cross-city generalizability (\cref{sec:q3}). Does pre-trained DGAE show robust when directly applied to a new city road network, and how does its accuracy compare to a model trained on data from the new city?




    \item \textbf{RQ4}: Ablation study (\cref{sec:q5}). Are key components in DGAE effective?

\end{itemize}

\subsection{Experimental setup}
\vspace{0.2cm}
\noindent \textbf{Dataset.} Three widely used open-source traffic speed datasets are evaluated, including METR-LA and PEMS-BAY from \cite{li2018dcrnn}, and PEMSD7(M) from \cite{yu2018STGCN}, the unit of the speed is mile per hour (mph). Descriptions for the datasets are as shown in \cref{tab:dataset}.
\vspace{0.2cm}

\noindent \textbf{Baselines.} Eight methods from existing research are selected as benchmarks. (1) \textbf{KNN}: K-Nearest neighbors, it performs interpolation based on the average of k-nearest neighbors. Neighbors are defined as different samples that exhibit "feature similarity". Since the speed feature of unobserved nodes are completely missing, we utilize auxiliary features such as latitude, longitude, and timestamps to facilitate interpolation for these unobserved nodes. The algorithm is implemented using the KNNImputer\footnote{\url{https://scikit-learn.org/stable/modules/generated/sklearn.impute.KNNImputer.html}} from the widely recognized machine learning library, scikit-learn. (2) \textbf{OKriging} \cite{saito2005OKriging}: Ordinary Kriging, which interpolates unseen location by using specific covariance and variogram functions to describe the spatial relationships of random variables at different locations. Gaussian variogram function is adopted is this work, and the algorithm is implemented based on Pykrige\footnote{\url{https://github.com/GeoStat-Framework/PyKrige}}, a well-developed kriging library. (3) \textbf{LETC} \cite{nie2023letc} Laplacian Enhanced low-rank Tensor Completion, which introduce graph Laplacian regulation and  temporal graph Fourier transform to enhance the tensor completion. (4) \textbf{IGNNK} \cite{wu2021IGNNK}: Inductive Graph Neural Networks for Kriging, which develops an inductive training procedure to estimate the unobserved locations state, the main block of IGNNK is consisted by three diffusion graph convolution layers using bidirectional random walk diffusion kernel. (5) \textbf{INCREASE} \cite{Zheng2023Increase}: Inductive graph representation learning model for spatio-temporal kriging, featuring GNN-based spatial aggregation and GRU-based temporal modeling to capture  heterogeneous spatial relations with various temporal patterns for interpolating state from unseen nodes. (6) \textbf{IAGCN} \cite{WEI2024IAGCN}: Inductive and Adaptive Graph Convolution Networks, which adopts an adaptive graph constructor to learn the spatial dependency for state estimation at sensor-free locations and introduces a predicted-based constraint task to enhance the estimation. (7) \textbf{KITS} \cite{xu2023kits}: Kriging model with Increment Training Strategy, which introduces additional fake nodes during training to mitigate the graph gap under the decremental training strategy. (8) \textbf{DEFP} \cite{rossi2022FP}: Dirichlet-Energy based Feature Propagation, a state-of-the-art graph diffusion-based model to estimate missing graph features.

\begin{table*}[t]

\caption{Performance comparison of DGAE against other baseline methods. The best result is highlighted in bold, while the second-best result is underlined. The "Improve" illustrates the percentage of relative performance improvement of DGAE compared to the most competitive baseline.}
\label{tab:q1}
\begin{tabularx}{\textwidth}{l*{9}{>{\centering\arraybackslash}X}}
\toprule
\multicolumn{1}{l}{\multirow{2}{*}{Method}} & \multicolumn{3}{c}{METR-LA}                       & \multicolumn{3}{c}{PEMS-BAY}                          & \multicolumn{3}{c}{PEMSD7(M)}    \\ \cmidrule(lr){2-10} 
\multicolumn{1}{c}{}                         & MAPE           & MAE           & RMSE            & MAPE          & MAE           & RMSE                  & MAPE              & MAE               & RMSE\\ \midrule
KNN                                          & 22.21          & 8.32          & 14.22           & 13.95         & 5.67          &  10.87                & 21.70             & 8.38              & 13.58    \\
OKriging                                     & 21.04          & 7.58          & 10.83           & 12.23         & 4.84          &  8.26                 & 19.96             & 7.40              & 13.86    \\
LETC                                         & 19.10          & 6.98          & 9.53            & 9.73          & 4.07          &  6.60                 & 22.01             & 7.47              & 10.50  \\
IGNNK                                        & 14.72          & 6.21          & 9.53            & {\ul 8.05}    & {\ul 3.59}    &  {\ul 5.93}           & 17.50             & {\ul 6.00}        & {\ul 9.19}   \\
INCREASE                                     & 14.98          & 6.06          & 9.24            & 8.35          & 3.64          &  6.33                 & 17.58             & 6.27              &  9.60   \\
IAGCN                                        & {\ul 14.23}    & {\ul 5.55}    & 9.17            & 8.34          & {\ul 3.59}    &  6.11                 & {\ul 17.10}       & 6.18              &  9.32   \\
KITS                                         & 14.95          & 5.58          & {\ul9.02}       & 8.46         & 3.57          &  6.45                 & 18.46             & 6.25              &  10.22  \\
DEFP                                         & 19.33          & 8.59          & 11.28           & 11.87         & 6.04          &  8.85                 & 19.70             & 8.15              & 11.12 \\
DEFP4D (ours)                                  & 15.55          & 6.06          & 9.04            & 8.76          & 4.03          &  6.70                 & 17.40             & 6.22              &  9.42   \\
DGAE (ours)                                  & \textbf{13.35} & \textbf{5.39} & \textbf{8.07}   & \textbf{7.32} & \textbf{3.36} &  \textbf{5.59}        & \textbf{15.61}    &\textbf{5.43}      &  \textbf{8.62}   \\ \midrule
Improve (\%)                                 & 6.18          & 2.88         & 10.53           & 9.07         & 6.41         &  5.73                & 8.71             & 9.50             &  6.20   \\ \bottomrule
\end{tabularx}
\end{table*}


\vspace{0.2cm}
\noindent \textbf{Metrics.} All the methods are evaluated by three commonly adopted metrics in time series modeling: Mean Absolute Percentage Error (MAPE), Mean Absolute Error (MAE) and Root Mean Squared Error (RMSE). They are defined as:

\begin{equation}
    \operatorname{MAPE} = \frac{1}{N*T}\sum_{i=1}^N\sum_{k=1}^T\left|\frac{\hat{\mathbf{x}}_{i,k}-\mathbf{x}_{i,k}}{\mathbf{x}_{i,k}}\right|\times 100\%
\end{equation}
\begin{equation}
    \operatorname{MAE} = \frac{1}{N*T}\sum_{i=1}^N\sum_{k=1}^T\left|\hat{\mathbf{x}}_{i,k}-\mathbf{x}_{i,k}\right|
\end{equation}
\begin{equation}
    \operatorname{RMSE} = \sqrt{\frac{1}{N*T}\sum_{i=1}^N\sum_{k=1}^T\left(\hat{\mathbf{x}}_{i,k}-\mathbf{x}_{i,k}\right)^2}
\end{equation}
where \(\hat{\mathbf{x}}_{i,k}\) denotes the estimation results of node \(i\) at time instance \(k\), and \(\hat{\mathbf{x}}_{i,k}\) denotes the corresponding ground truth. \(N\) and \(T\) are the total number of unobserved nodes and the length of the time window under consideration within each estimation, respectively. It is important to note that the original missing values in the dataset are excluded from the evaluation.

\vspace{0.2cm}
\noindent \textbf{Implementation.} The proposed DGAE is implemented with PyTorch 2.2.1 on an NVIDIA RTX 4090 GPU. We update the estimation results every hour, and each round of estimation includes 12-step time instances. The time step length is set as 5 min. Mean square error is adopted as the loss function. The dataset is divided into training and testing sets in a 7:3 ratio. The validation set is kept the same as the training set and used for early stopping with patience set to 10. Unless explicitly stated otherwise, following previous works \cite{wu2021IGNNK,Zheng2023Increase,WEI2024IAGCN}, we randomly select 25\% of the available sensors as Virtual Sensors (VS), whose data are only used for evaluation. To simulate the inductive setting, the data collected by VS, as well as their locations, are not be disclosed during the training and validation processes. The remaining sensors are referred to as Available Sensors (AS) for brevity.

\subsection{Overall state estimation performance (RQ1)} \label{sec:q1}

\cref{tab:q1} presents the performance comparison of proposed DEFP4D and DGAE model against other baseline methods. It is evident that DGAE consistently outperforms all baseline methods across different metrics on three datasets, especially in METR-LA. The improvement achieved by DGAE against the most competitive baseline averaged 7.25\% and reached up to 10.53\%. Traditional methods exhibit clear limitations in this task. Due to the lack of available traffic features for unobserved nodes, the KNN algorithm is limited to relying solely on timestamps and geographic coordinates for neighbor selection. Unfortunately, spatial proximity based on geographic coordinates is ill-suited for the intricate, non-Euclidean structures of traffic networks, leading to substantial inaccuracies in KNN’s estimation. In contrast, OKriging utilizes the speed features of observed nodes to fit functions that describe the covariance and spatial relationships among different node states, enabling interpolation for unobserved nodes. However, OKriging encounters challenges in capturing temporal dependencies, and its simplified variogram functions are insufficient for non-Euclidean traffic networks, subsequently limiting its estimation accuracy. In comparison, IGNNK fully harnesses the capabilities of bidirectional graph diffusion neural networks, achieving superior performance in several experiments among baselines. Nonetheless, IGNNK adopts a zero-filling strategy for missing node data, which introduces additional noise into the learning process, leading to lower performance metrics compared to our proposed DGAE. INCREASE model introduces GRU-based temporal dependency modeling, enabling it to outperform IGNNK in METR-LA dataset. However, it should be noted that many existing studies have confirmed that MLP is sufficient for modeling time dependence in series-related task \cite{Shao2022STID,wang2024timemixer}, which may be the reason why INCREASE does not outperform IGNNK in another two datasets. Different from IGNNK and INCREASE, IAGCN enhances performance through adaptive spatial dependency modeling and a traffic prediction-oriented auxiliary task. However, its reliance on zero-filling for unobserved nodes similarly introduces noise into the model, potentially limiting its effectiveness. Notably, DEFP demonstrates the fundamental effectiveness of Dirichlet energy minimization by outperforming OKriging through a simple feature propagation mechanism, though its undirected graph design limits its applicability in asymmetric traffic networks. In this paper, we propose DEFP4D as a directed graph adaptation of Dirichlet energy-based feature propagation, which also serves as a lightweight version of our DGAE model without deep learning components. DEFP4D surprisingly shows competitive performance, ranking second in RMSE on METR-LA dataset. These success validate the potential of Dirichlet energy for TSE tasks, though there remains room for improvement in capturing complex spatiotemporal dynamics in traffic networks.

\begin{figure*}
    \centering
    \includegraphics[width=0.9\linewidth]{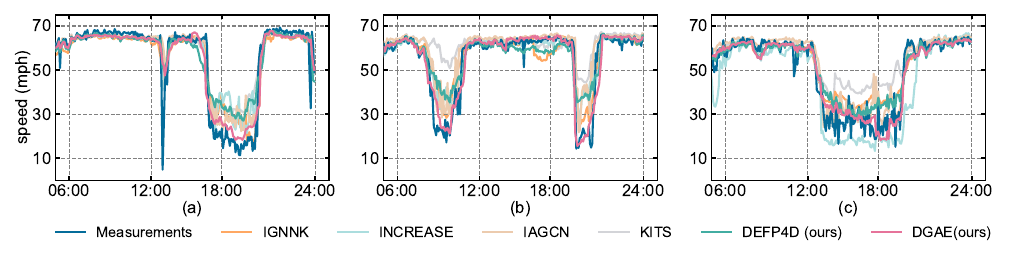}
    \vspace{-7mm}
    \caption{State estimation results for METR-LA: (a) Sensor \#1, (b) Sensor \#70, (c) Sensor \#131.}
    \vspace{-12pt}
    \label{fig:q1}
\end{figure*}
\begin{figure*}
    \centering
    \includegraphics[width=0.9\linewidth]{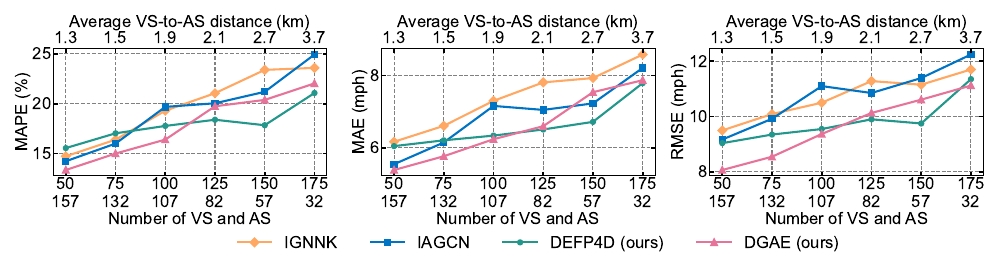}
    \vspace{-7mm}
    \caption{Model performance with different sensor deployment density on METR-LA dataset. The first row of x-ticks labels indicates the number of VS, while the second row shows the number of VS.}
    \vspace{-12pt}
    \label{fig:q2}
\end{figure*}
Given the paramount importance of congestion evolution in traffic management tasks, the data from METR-LA that exhibit significant congestion characteristics are selected for further analysis. As shown in \cref{fig:q1}, it illustrates the comparison between the measured and estimated values of three sensors (\#1,\#70,\#131) during a day from 6:00 to 24:00. It is clear that our DGAE model can successfully captures the onset and evolution of congestion at unobserved nodes, achieving the best performance against the baselines. For example, at the location where sensor \#131 is deployed, INCREASE tends to overestimate the congestion level, while the other models underestimate it. However, our DGAE model, represented by the pink curve, accurately and smoothly tracks the congestion speed measurement curve, demonstrating strong state estimation capabilities.

\subsection{Impact of sensor deployment density (RQ2)} \label{sec:q2}

To assess the effect of sensor deployment density on model performance, we systematically increase the number of VS from 50 to 175 within the METR-LA dataset. We compare our proposed DGAE and DEFP4D with the two most competitive baselines, IGNNK and IAGCN. The performance trends corresponding to these changes are depicted in \cref{fig:q2}. Additionally, we introduce a new metric, the Average VS-to-AS Distance (\(D_{v2a}\)). The \(D_{v2a}\) distance increases with the number of VS, reflecting the model's spatial extrapolation capabilities, as indicated at the top of the x-axis in \cref{fig:q2}. Its definition is as follows:
\begin{equation}
    d_{min}^i = \min_{j:j \in \text{AS, } i \neq j}d_{i,j}
\end{equation}
\begin{equation}
    D_{v2a} = \operatorname{avg}_{i \in \text{VS}} d_{min}^i
\end{equation}

As shown in \cref{fig:q2}, when the number of VS is 100 or fewer, by which case the average distance (\(D_{v2a}\)) from VS to AS is no more than 1.9 km, our proposed DGAE consistently outperforms all other methods. Specifically, when the number of VS is 100 and \(D_{v2a}\) is 1.9 km, the estimation error of DGAE compares favorably with that of IAGCN and IGNNK, which have a number of VS equal to 50 and \(D_{v2a}\) equal to 1.3 km. This result indicates that, under the same performance requirements, DGAE can rely on fewer sensors, effectively reducing the deployment cost of sensors.

When the number of VS exceeds 100, the performance of the proposed DEFP4D begins to surpass that of DGAE, fully overtaking it and other baseline methods once the number of VS reaches 125, corresponding to an \(D_{v2a}\) of 2.1 km. This indicates that when sensors are insufficient, i.e., the deployment is too sparse, complex data-driven methods tend to over-fit the limited observations and struggle to produce reliable results. In such scenarios, the lightweight DEFP4D emerges as the optimal choice, highlighting the need for careful consideration of model selection based on the density of deployed sensors in practical applications. Furthermore, it is important to note that DGAE masks a portion of AS as VS during the training procedure, further contributing to the sparsity of available detectors. Such limitations underscore the necessity of developing alternative training strategies to alleviate the sparsity problem so as to improve model performance.

\begin{table}[t]
\caption{Cross-city generalization performance evaluation. The model was trained exclusively on the Los Angeles (METR-LA) dataset and directly tested on the Bay Area (PEMS-BAY) dataset without any fine-tuning or adaptation.}
\label{tab:q3}
\begin{tabularx}{\columnwidth}{lXXX}
\toprule
Method & MAPE  & MAE  & RMSE \\ \midrule
IGNNK & 8.57 & 4.01 & 6.63 \\
KITS & {\ul8.50} & {\ul3.65} & {\ul6.47} \\ 
DGAE & \textbf{8.11} & \textbf{3.63} & \textbf{6.09} \\  \midrule
DGAE (retrained)   & 7.32 & 3.36 & 5.59 \\
\bottomrule
\end{tabularx}
\end{table}

\begin{figure*}[t]
    \centering
    \includegraphics[width=0.9\linewidth]{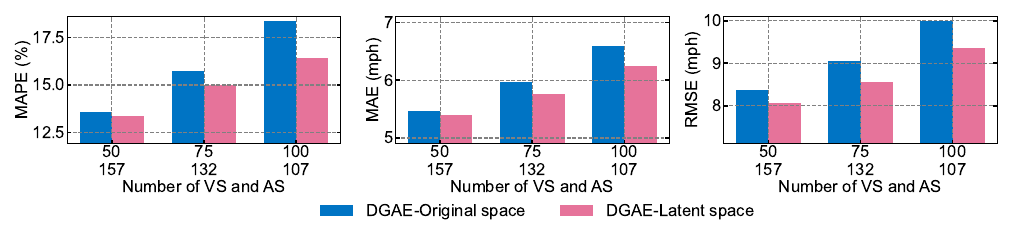}
    \vspace{-7mm}
    \caption{The effectiveness of DEFP4D in latent space representation. We compare our proposed DGAE with a variant that applied DEFP4D in the original feature space. The numbers 50, 75, and 100 indicate the number of unobserved nodes in the traffic network of METR-LA dataset.}
    \vspace{-12pt}    
    \label{fig:q6-1}
\end{figure*}

\subsection{Cross-city generalizability (RQ3)} \label{sec:q3}
In this section, we examine the model's ability to generalize across different cities, focusing exclusively on direct transfer performance without any fine-tuning procedures. Building models with excellent cross-city generalization capabilities offers two significant advantages: On one hand, it substantially reduces deployment costs since the pre-trained model requires only minimal fine-tuning in target cities; On the other hand, for cities constrained by privacy protection policies that can only access real-time stream data and are not allowed to store historical records, a well-generalized model trained on other cities provides not only direct prediction services but also establishes a solid basis for implementing online learning algorithms such as continual learning. 

Therefore, we conduct experiments by training our model on the Los Angeles (METR-LA) dataset and evaluating its transfer performance on the Bay Area (PEMS-BAY) dataset. We select IGNNK and KITS as baseline models as they are the two most competitive models on METR-LA among methods that support direct transfer without retraining. As shown in \cref{tab:q3}, our model maintains its superior performance after transferring to PEMS-BAY, demonstrating its practical value and great potential as a foundation model for transfer learning or continual learning scenarios. Notably, comparing with the results in \cref{tab:q1}, we observe varying degrees of performance degradation across models. KITS exhibits the smallest performance drop, which might be attributed to its unique incremental graph learning strategy. This observation provides valuable insights for future research directions.


\subsection{Ablation study (RQ4)} \label{sec:q5}


To evaluate the effectiveness of the key components in DGAE, we conduct the following ablation experiments on METR-LA dataset.

1) To investigate the benefit of applying DEFP4D in latent space, we conduct ablation studies by reverting DGAE to the classic graph auto-encoder defined in \cref{eq:tge} and \cref{eq:tgd}. To deal with unobserved nodes, we apply DEFP4D in the original feature space to get their initial values. Based on the results in \cref{fig:q6-1}, DGAE-Latent space shows better performance than DGAE-Original space across all scenarios. The advantage becomes more pronounced as the proportion of unobserved nodes increases. This confirms the effectiveness of applying DEFP4D in the latent space for speed estimation.

2) To validate the necessity of decoupling congestion and free-flow signal propagation, we replace the decoupled bidirectional diffusion with coupled one and unidirectional diffusion. Reference \cref{tab:q6-2}, we can find that one-way diffusion in the direction of congestion propagation is not much different from that in the direction of free-flow propagation, and the effect lags far behind the bidirectional diffusion, which proves the effectiveness of bidirectional diffusion. In addition, decoupling the diffusion signal improves the model performance.

3) To assess the necessity of graph diffusion operations, we replace graph diffusion in the encoder and decoder with MLP. As shown in \cref{tab:q6-4}, in both encoder and decoder, graph diffusion improves the estimation performance, which implies the necessity of mixing the features of the observed nodes before and after embeddings of new nodes are generated.



\begin{table}[t]
\caption{The effectiveness analysis of decoupling traffic flow propagation into congestion and free flow signal. DGAE-A and DGAE-B are the DGAE with only free flow and congestion propagation direction, respectively. DGAE-AB does not decouple bidirectional propagation, i.e., the transition matrix derived in \cref{eq:update} is maintained.}
\label{tab:q6-2}
\begin{tabularx}{\columnwidth}{XXXX}
\toprule
Method & MAPE  & MAE  & RMSE \\ \midrule
DGAE-A & 16.27 & 6.05 & 9.04 \\
DGAE-B & 16.35 & 6.38 & 9.50 \\
DGAE-AB & {13.91} & {5.72} & {8.55} \\ \midrule
DGAE   & \textbf{13.35} & \textbf{5.39} & \textbf{8.07} \\ \bottomrule
\end{tabularx}
\end{table}

\begin{table}[t]
\caption{The effectiveness analysis of graph diffusion-based encoder and decoder. DGAE-MLPEn and DGAE-MLPDe replace the graph diffusion in the encoder and decoder with multilayer perceptron layer, respectively.}
\label{tab:q6-4}
\begin{tabularx}{\columnwidth}{XXXX}
\toprule
Method   & MAPE  & MAE  & RMSE \\ \midrule
DGAE-MLPEn & 14.14 & 5.67 & 8.53 \\
DGAE-MLPDe & 13.95 & 5.66 & 8.50  \\ \midrule
DGAE    & \textbf{13.35} & \textbf{5.39} & \textbf{8.07} \\ \bottomrule
\end{tabularx}
\end{table}

\begin{figure*}[t]
    \centering
    \includegraphics[width=1.0\linewidth]{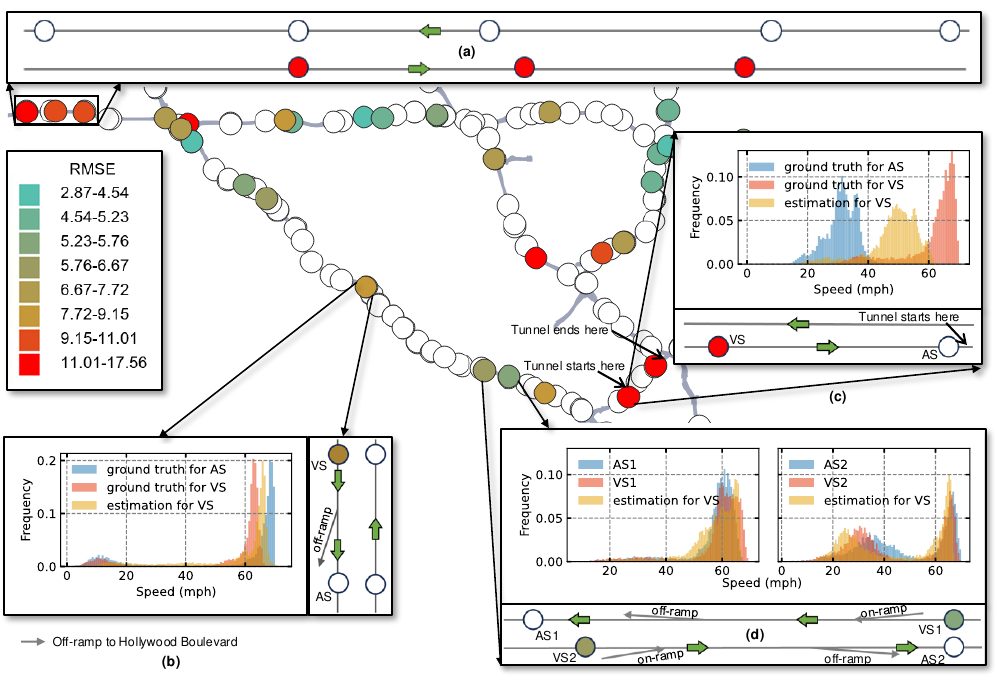}
    \vspace{-8mm}
    \caption{A map view of estimation error distribution. The uncolored and colored circles indicate AS and VS, respectively, where different colors indicate different levels of RMSE of the estimates. Subplots (a)-(d) show zoomed-in views of sensor spatial distribution and statistical distribution of data in four selected scenarios.}
    \vspace{-12pt}
    \label{fig:error-heat}
\end{figure*}

\section{Discussion} \label{sec:diss}
While our experimental results demonstrate promising overall performance, further analysis of the prediction errors reveals some intriguing patterns that warrant further investigation. Through examining the spatial distribution of estimation errors, we observe that the model's performance varies across different locations within the traffic network. This spatial heterogeneity of prediction accuracy, though potentially crucial for practical applications, remains an open question for current research. To gain deeper insights from this phenomenon, we visualize and analyze the spatial distribution of estimation errors and select several representative scenarios for detailed discussion, as shown in \cref{fig:error-heat}. In this visualization, uncolored and colored circles represent the locations of AS and VS respectively, where the color intensity of VS indicates the magnitude of estimation error (RMSE). Four representative scenarios (a-d) are selected for detailed analysis, each accompanied by a detailed schematic highlighting the spatial relationship between VS and nearby AS, along with comparative speed distributions of the corresponding detection points.

In scenario (a), no AS is available in the same direction as VS, with all nearby AS located in the opposite direction. This lack of effective conditional information evidently impacts the model's estimation accuracy at VS locations. This observation emphasizes the need to understand the boundaries of model capabilities and quantify estimation uncertainty, preventing the model from providing potentially unreliable estimations beyond its effective scope.

In scenario (b), an off-ramp between the VS and AS locations connects to Hollywood Boulevard, a major arterial road with substantial exit demand. The significant exit volume leads to early deceleration behavior that notably affects mainline traffic conditions. As evidenced by the speed distributions, the free-flow speed at VS upstream of the off-ramp is considerably lower than that at AS, resulting in larger estimation errors. Although our model's estimated distribution shows an effort to capture this speed reduction by shifting towards the VS pattern rather than simply replicating the AS distribution, the complex ramp-induced traffic dynamics still pose significant challenges for accurate estimation. This highlights the critical role of ramp effects in traffic state estimation.

In scenario (c), AS near VS is situated at the tunnel entrance, where Google Street View reveals a posted speed limit of 40 mph, consistent with the observed AS speed distribution. Despite our model's estimated distribution showing a trend towards the lower speeds observed at VS rather than fully adhering to the AS pattern, the infrastructure-induced speed heterogeneity still leads to considerable estimation errors. This case reveals the importance of modeling traffic flow heterogeneity induced by infrastructure characteristics, an aspect that has received limited attention in current data-driven approaches, including our model

In scenario (d), although on- and off-ramps are present between VS and AS, their connections to lower-class roads result in minimal flow exchange with the mainline. Additionally, the auxiliary lane extending between these closely-spaced ramps effectively mitigates their impact on mainline traffic. This is reflected in the similar speed distributions between VS and AS, leading to acceptable estimation errors. When considered alongside scenario (b), this suggests that ramp effects exhibit spatial-temporal heterogeneity, providing valuable insights for future model development.

Based on these analyses, we identify several promising research directions for enhancing traffic state estimation models. First, there is a clear need to calibrate model capability boundaries and incorporate uncertainty quantification, particularly in scenarios where limited or unrepresentative conditional information is available. Second, the contrasting observations between different ramp scenarios highlight the importance of characterizing spatially heterogeneous ramp effects, which could significantly improve estimation accuracy in complex urban environments. Finally, our findings underscore the necessity of explicitly modeling infrastructure-dependent traffic flow patterns, as infrastructure characteristics can substantially influence traffic behavior in ways that current data-driven approaches may not fully capture. These directions, while challenging, represent crucial steps toward developing more robust and reliable traffic state estimation models.


\section{Conclusion} \label{sec:con}

Previous studies have demonstrated the effectiveness of inductive graph neural networks in traffic state estimation tasks, showing remarkable operational flexibility in seamlessly adapting to new unobserved locations of interest, varying sensor configurations due to inevitable retirement and new installations, and cross-city applications. However, existing methods typically adopt zero-filling strategies when pre-processing unobserved node features to facilitate rapid GNN deployment, inevitably introducing substantial errors in node embedding learning. This work first proposes the Dirichlet Energy-based Feature Propagation algorithm for Directed graphs (DEFP4D) and integrates it with a graph auto-encoder to form the DGAE model. DGAE initially focuses on learning node embeddings for observed nodes, then generates embeddings for unobserved nodes using DEFP4D in the latent space, effectively avoiding error propagation from zero-filling. Leveraging the flexibility of the graph auto-encoder, we decouple the propagation of congested and free-flow signals, better characterizing traffic flow dynamics and significantly enhancing model performance.

Extensive experiments on three open-source large-scale traffic datasets demonstrate that our method surpasses existing approaches by an average margin of 7.25\% in traffic speed estimation, with improvements reaching up to 10.53\%. As the proportion of unobserved nodes progressively increases, DGAE consistently maintains its superior performance over existing methods, achieving comparable estimation performance with 14.5\% fewer sensors than the most competitive baseline. The method's effectiveness extends to cross-city transfer tasks, where DGAE maintains its performance advantages, validating its generalization capabilities. Additionally, DEFP4D, the lightweight variant of DGAE without the auto-encoder structure, demonstrates competitive performance by surpassing existing SOTA graph representation learning methods in some scenarios. When the unobserved rate exceeds 50\%, DEFP4D even outperforms its more complex counterpart DGAE, suggesting that simpler propagation mechanisms may exhibit superior robustness in extremely sparse observation scenarios.

Despite these superior performance achievements, several promising research directions remain to be explored: 1) Uncertainty quantification and calibration for model predictions, particularly in scenarios with limited conditional information; 2) Integration of infrastructure characteristics, including ramp effects heterogeneity and speed limits information; 3) Multi-source data fusion under privacy preservation constraints, considering the sensitive nature of transportation data from different stakeholders.

\section*{Data availability}
The data that support the findings of this study are publicly available. They can be accessed at \href{https://www.example-database.com}{Performance Measurement System (PeMS)}.


%





\ifCLASSOPTIONcaptionsoff
  \newpage
\fi





\bibliographystyle{IEEEtran}
\bibliography{IEEEabrv,Bibliography}
%

\begin{IEEEbiography}[{\includegraphics[width=1in,height=1.25in,clip,keepaspectratio]{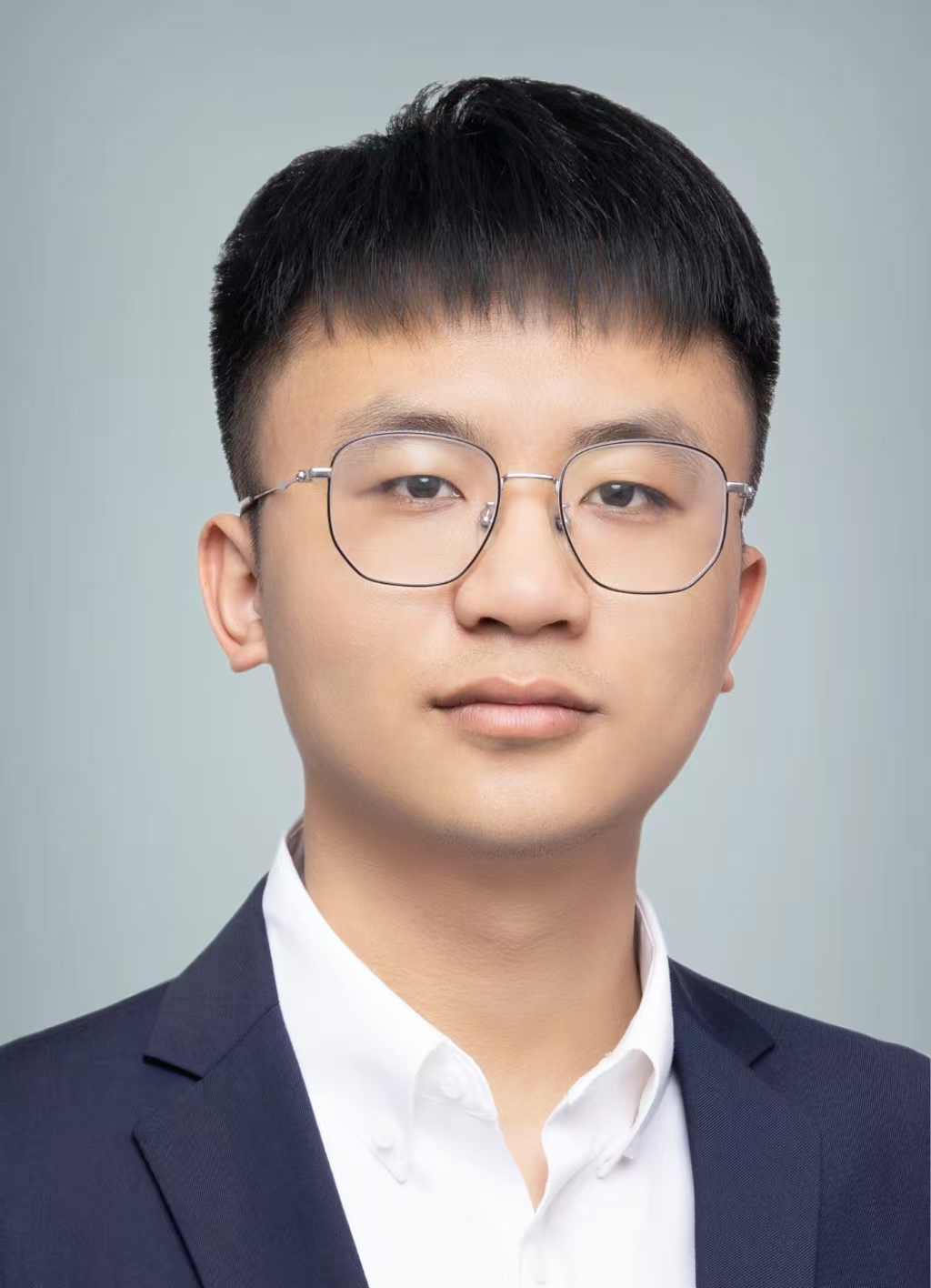}}]{Qishen Zhou} received the B.S. degree in transportation engineering from the Huazhong University of Science and Technology in 2020. He is currently pursuing the Ph.D. degree with the College of Civil Engineering and Architecture, Zhejiang University. His research interests include traffic state estimation and traffic control based on connected and autonomous vehicles.
\end{IEEEbiography}

\begin{IEEEbiography}
    [{\includegraphics[width=1in,height=1.25in,clip,keepaspectratio]{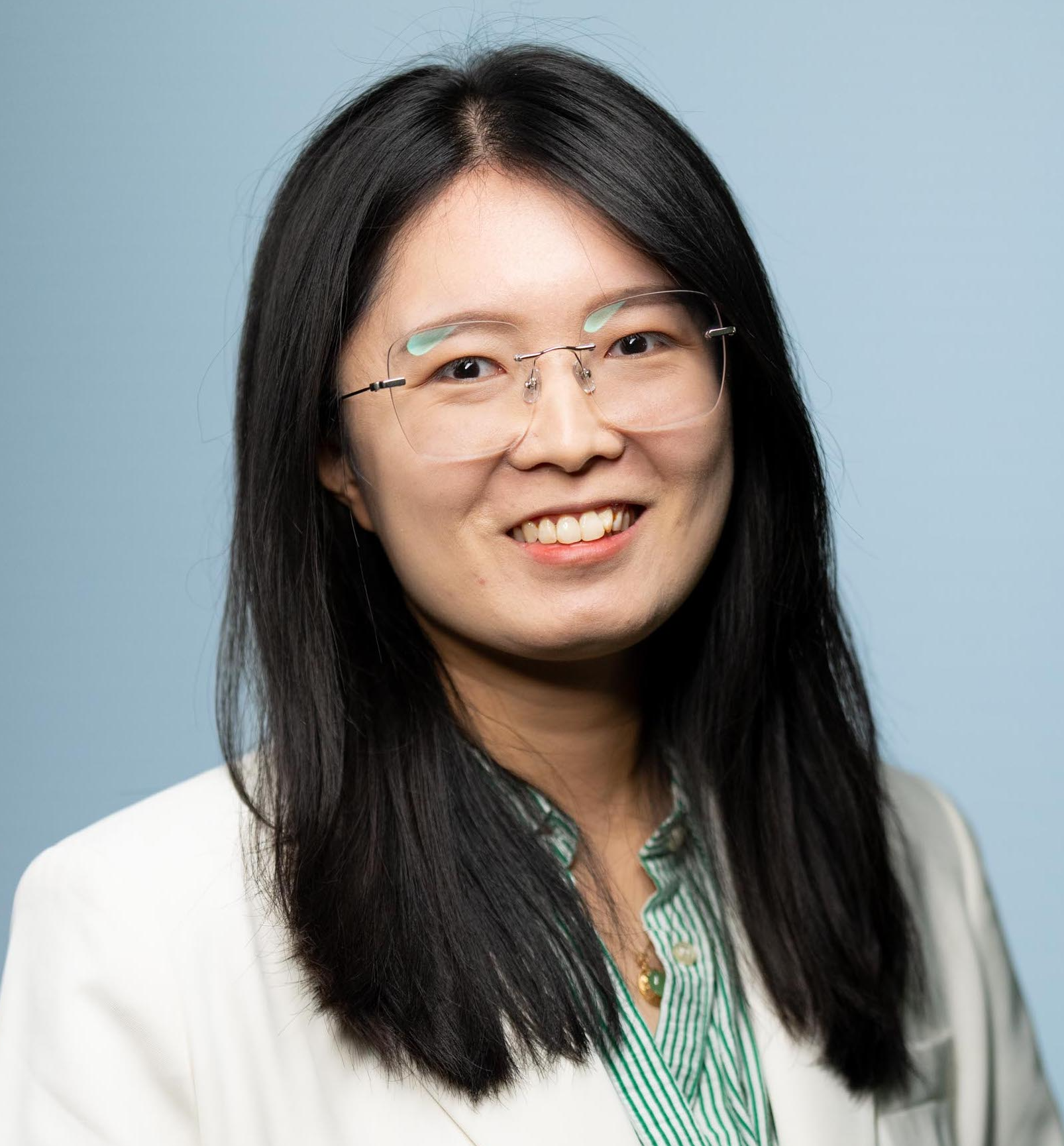}}]{Yifan Zhang} received the B.Eng. degree in telecommunication from Xidian University, Xi'an, China, in 2015, and the M.Sc. degree in computer science from ENSIIE, \'{E}vry, France, in 2017. She received the Ph.D degree in computer science from City University of Hong Kong, in 2022. She is currently an Assistant Professor in the Department of Computer Science at City University of Hong Kong (Dongguan). Her research interests include spatial-temporal modeling, motion planning, microscopic traffic flow modeling, autonomous driving, and intelligent transportation systems.
\end{IEEEbiography}

\begin{IEEEbiography}[{\includegraphics[width=1in,height=1.25in,clip,keepaspectratio]{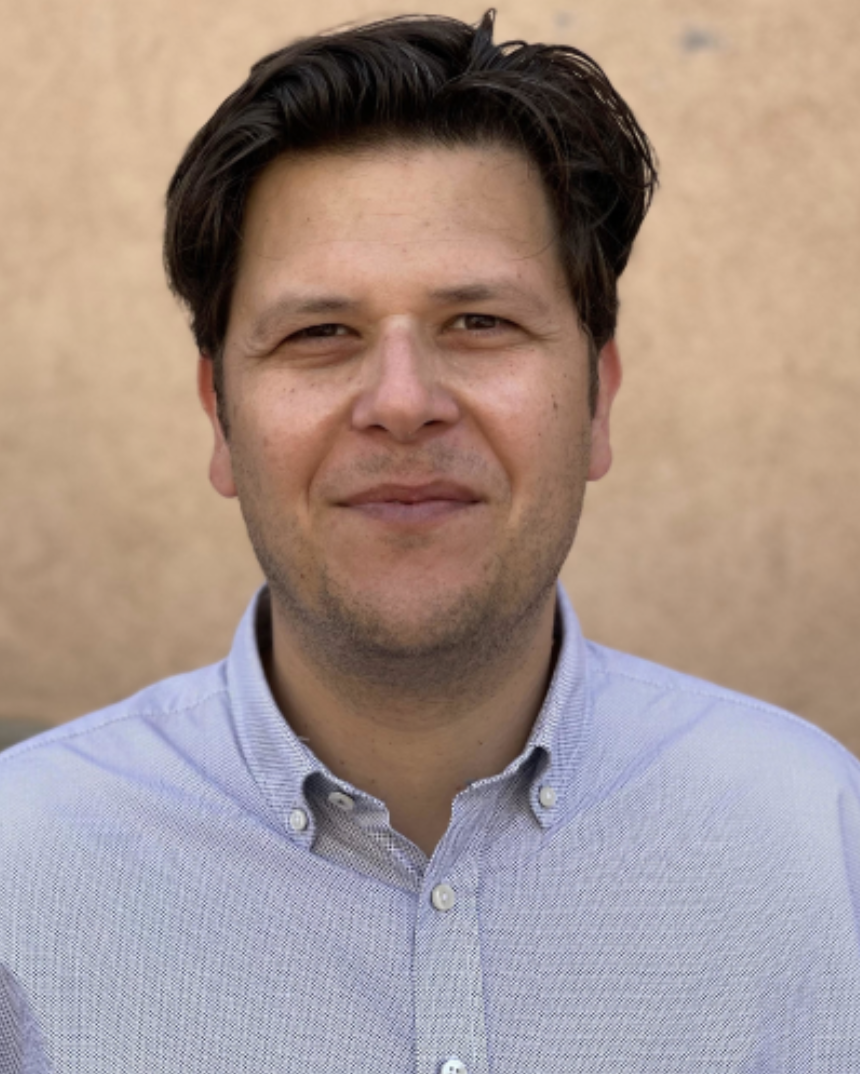}}]{Michail A. Makridis} (Member, IEEE) received
the Ph.D. degree in computer vision from the
Democritus University of Thrace, Greece. He is
currently a Senior Research Scientist and the Deputy
Director of the Traffic Engineering Group, Swiss
Federal Institute of Technology (ETH) Zürich.
Before, he was the scientific responsible for the
Traffic Modeling Group, Sustainable Transport Unit,
Joint Research Centre (JRC), European Commission
(EC). His research interests include traffic management, simulation, and control for future intelligent
transportation systems in the presence of connected and automated vehicles.
In 2022, he received the JRC Annual Awards on Excellence in Research from
the EC.
\end{IEEEbiography}

\begin{IEEEbiography}[{\includegraphics[width=1in,height=1.25in,clip,keepaspectratio]{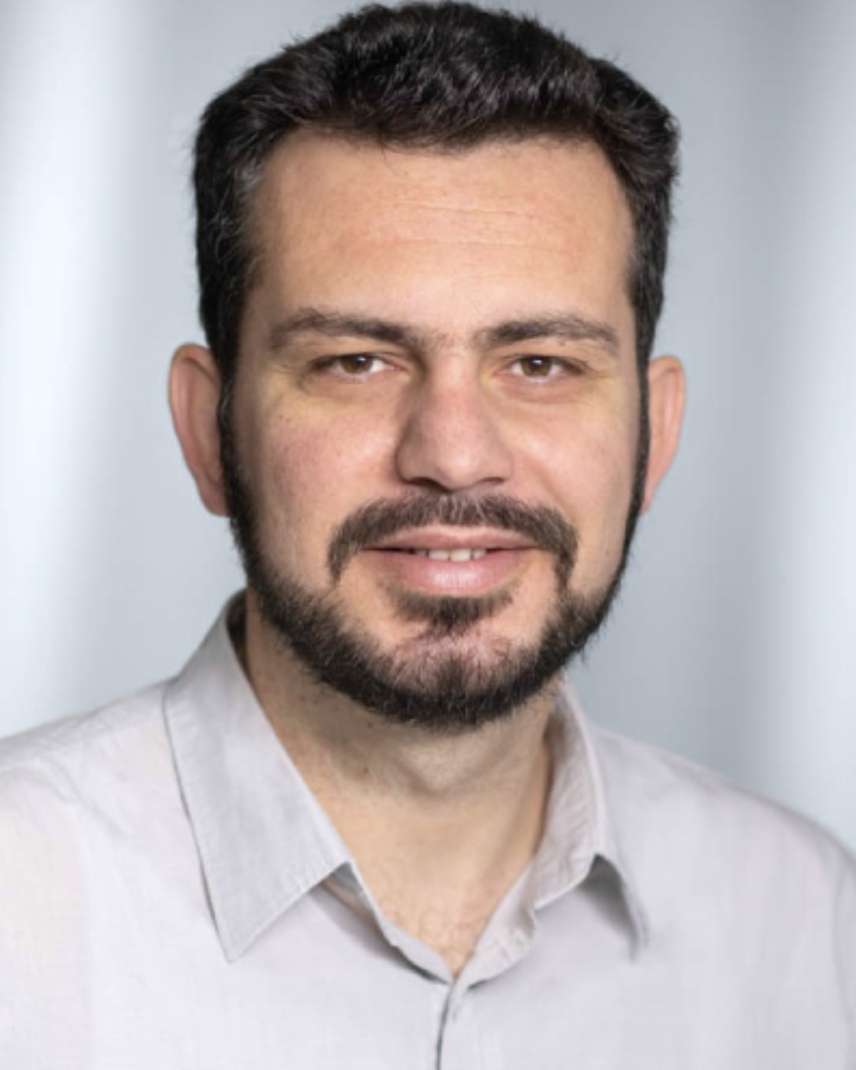}}]{Anastasios Kouvelas} (Senior Member, IEEE)
received the Diploma, M.Sc., and Ph.D. degrees
from the Department of Production and Management Engineering (Operations Research), Technical
University of Crete, Greece, in 2004, 2006, and
2011, respectively, specializing in modeling, control,
and optimization of large-scale transport systems.
He has been the Director of the Research Group
Traffic Engineering and Control, Institute for Transport Planning and Systems (IVT), Department of
Civil, Environmental and Geomatic Engineering,
Swiss Federal Institute of Technology (ETH) Zürich, since August 2018.
Prior to joining IVT, he was a Research Scientist with the Urban Transport Systems Laboratory (LUTS), EPFL (2014–2018), and a Post-Doctoral
Fellow with Partners for Advanced Transportation Technology (PATH),
University of California at Berkeley (2012–2014). Before this, he was
appointed as an Adjunct Professor with the Technical University of Crete
in 2011 and a Research Associate with the Centre for Research and Technology Hellas (CERTH), Information Technologies Institute, Thessaloniki,
Greece (2011–2012). In 2009, he was a Doctoral Visiting Scholar with the
Center for Advanced Transportation Technologies (CATT), Viterbi School of
Engineering, Department of Electrical Engineering, University of Southern
California, Los Angeles. He has been awarded with the 2012 Best IEEE
ITS PhD Dissertation Award from IEEE Intelligent Transportation Systems
Society. He has been a member of the Transportation Research Board (TRB)
Standing Committee (AHB15) on Intelligent Transportation Systems since
2019. He has been serving as an Associate Editor for IET Intelligent Transport
Systems journal since 2017. He has guest-edited several special issues in
transportation journals.
\end{IEEEbiography}

\begin{IEEEbiography}[{\includegraphics[width=1in,height=1.25in,clip,keepaspectratio]{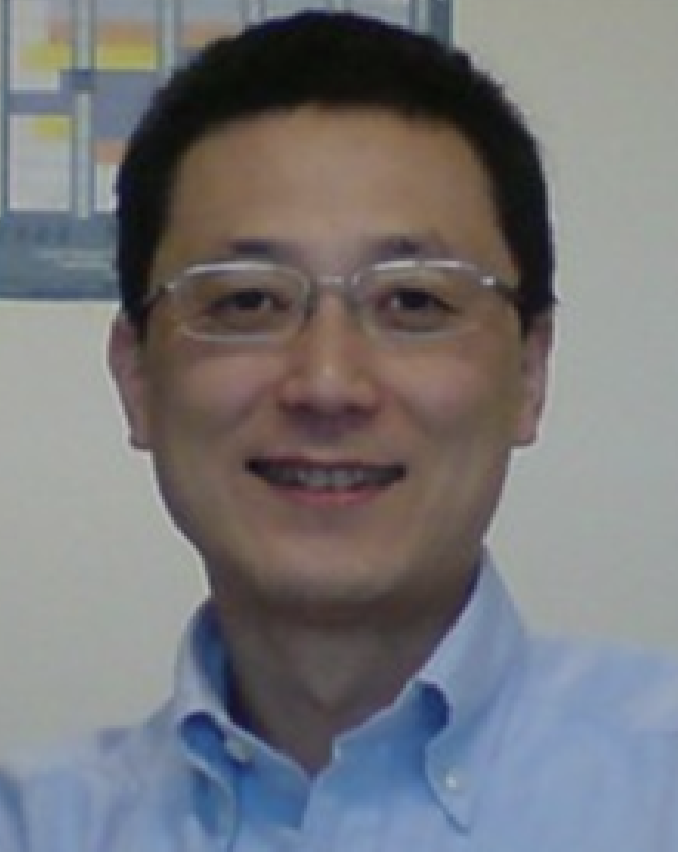}}] {Yibing Wang} (Senior Member, IEEE) received the
B.Sc. degree in electronics and computer engineering from Sichuan University, Chengdu, China, the
M.Eng. degree in automatic control engineering
from Chongqing University, Chongqing, China, and
the Ph.D. degree in control theory and applications
from Tsinghua University, Beijing, China. He was
a Post-Doctoral Researcher/Research Fellow/Senior
Research Fellow with the Dynamic Systems and
Simulation Laboratory, Department of Production
Engineering and Management, Technical University of Crete, Greece, from 1999 to 2007. He was a Senior Lecturer
with the Department of Civil Engineering, Monash University, Australia,
from 2007 to 2013. Since 2013, he has been a Full Professor with the
Institute of Intelligent Transportation Systems, Zhejiang University, China. His
research interests include road traffic flow modeling, surveillance, and control,
connected automated vehicles (CAVs). His research has been supported by the
European Commission, the Discovery Program of the Australian Research
Council, the National Natural Science Foundation of China, the National
Key Research and Development Program of China, and the Provincial Key
Research and Development Program of Zhejiang. He has published more
than 80 refereed papers in scientific journals. He serves as a Senior Editor
for IEEE TRANSACTIONS ON INTELLIGENT TRANSPORTATION SYSTEMS
and an Associate Editor for Transportation Research Part C: Emerging
Technologies. He was elected by the Zhejiang Qian Ren Program in 2012, and
recognized as the Leading Academic Talent (First Batch) in International Road
Transport Science and Technology by the China Highway and Transportation
Society in 2022. He and his students were the recipients of the First Best Paper
Award at the 23rd IEEE International Conference on Intelligent Transportation
Systems in 2020.
\end{IEEEbiography}

\begin{IEEEbiography}
[{\includegraphics[width=1in,height=1.25in,clip,keepaspectratio]{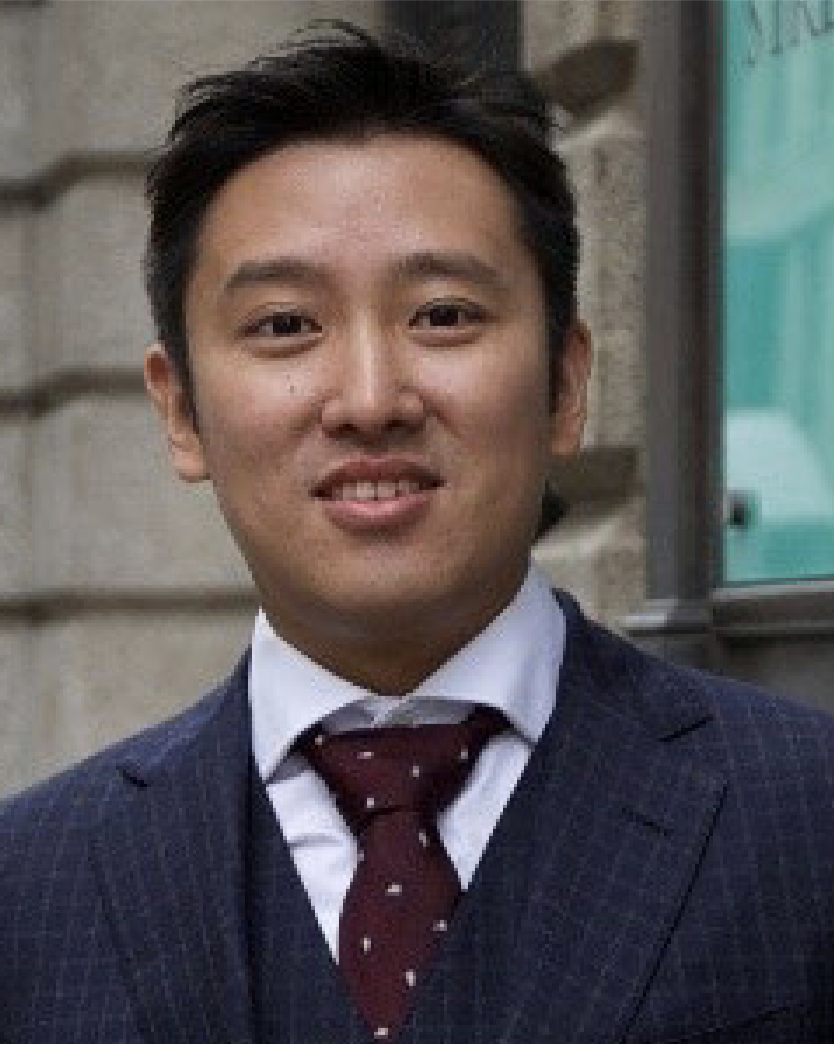}}]{Simon Hu} (Senior Member, IEEE) received the M.Sc. and
Ph.D. degrees in transportation engineering from
Imperial College London, U.K., in 2007 and 2011,
respectively. He is currently an Assistant Professor
with the Zhejiang University–University of Illinois
Urbana–Champaign Institute, Zhejiang University, China. He is also an Adjunct Professor at
the University of Illinois Urbana–Champaign (UIUC)
and an Honorary Research Fellow with Imperial
College London (ICL). He has undertaken more than
ten government-sponsored research projects. He has
published over 80 international journal articles and conference papers. His
research interests include intelligent transportation systems, connected and
automated vehicles, smart mobility, and the environmental impact of transportation systems. He is a member of the Chartered Institute of Highways
and Transport Engineering (CIHT). He serves as an Associate Editor for
IEEE TRANSACTIONS ON INTELLIGENT VEHICLES and the IEEE Intelligent Transportation Systems Conference. He also serves as a Scientific Committee member for the European Association for Research in Transportation (hEART) and a member of the Standing Committee on Alternative Fuels and Technologies (AMS40) at the Transportation Research Board (TRB).
\end{IEEEbiography}





\vfill


\end{document}